\documentclass[11pt]{article}

\usepackage[utf8]{inputenc}
\usepackage[T1]{fontenc}
\usepackage{hyphenat}
\usepackage{xspace}
\usepackage{amsmath}
\usepackage{amsfonts}
\usepackage{hyperref}
\usepackage{url}
\usepackage{booktabs}
\usepackage{multirow}
\usepackage{makecell}
\usepackage{caption}
\usepackage{minibox}
\usepackage{bbm}
\usepackage{graphicx}
\usepackage{balance}
\usepackage{mathtools}
\usepackage{color}
\usepackage{marvosym}
\usepackage{ifthen}
\usepackage{textcomp}
\usepackage{enumitem}
\usepackage{verbatim}
\usepackage{algorithm}
\usepackage{numprint}

\usepackage{amsthm}
\theoremstyle{plain}
\newtheorem{theorem}{Theorem}[section]
\newtheorem{lemma}[theorem]{Lemma}
\newtheorem{proposition}{Proposition}[section]
\newtheorem{corollary}{Corollary}[theorem]
\newtheorem{definition}{Definition}[section] %
\newtheorem{example}{Example}[section]%

\newcommand{\chatoDisplayMode}[1]{#1}

\definecolor{MyRed}{rgb}{0.6,0.0,0.0}
\definecolor{MyBlack}{rgb}{0.1,0.1,0.1}
\newcommand{\inred}[1]{{\color{MyRed}\sf\textbf{\textsc{#1}}}}
\newcommand{\frameit}[2]{
  \begin{center}
  {\color{MyRed}
  \framebox[.9\columnwidth][l]{
    \begin{minipage}{.85\columnwidth}
    \inred{#1}: {\sf\color{MyBlack}#2}
    \end{minipage}
  }\\
  }
  \end{center}
}

\newcommand{\note}[2][]{\chatoDisplayMode{\def\@tmpsig{#1}\frameit{{\Pointinghand} Note}{#2\ifx \@tmpsig \@empty \else \mbox{ --\em #1}\fi}}}
\newcommand{\todo}[2][]{\chatoDisplayMode{\def\@tmpsig{#1}\frameit{{\Writinghand} To-do}{#2\ifx \@tmpsig \@empty \else \mbox{ --\em #1}\fi}}}

\newcommand{\abbrevStyle}[1]{#1}

\usepackage{amssymb}%
\usepackage{pifont}%

\newcommand{\ie}{\abbrevStyle{i.e.}\xspace}

\newcommand{\Tabref}[1]{Table~\ref{#1}}

\newcommand{\Figref}[1]{Fig.~\ref{#1}}

\newcommand{\Thmref}[1]{Thm.~\ref{#1}}

\newcommand{\xhdr}[1]{\vspace{1.7mm}\noindent{{\bf #1.}}}

\newcommand{\xhdrNoPeriod}[1]{\vspace{1.7mm}\noindent{{\bf #1}}}

\newcommand{\denselist}{ \itemsep -2pt\topsep-10pt\partopsep-10pt }

\newcommand{\textcite}[1]{\citeauthor{#1} \shortcite{#1}}

\newcommand{\hide}[1]{}

\makeatletter
\newcommand{\iffont}[2]{\ifthenelse{\equal{\f@family}{#1}}{#2}{}}
\makeatother

  \usepackage{mathptmx}

  \DeclareSymbolFont{greek}{OML}{cmm}{m}{n}
  \DeclareMathSymbol{\alpha}{\mathalpha}{greek}{"0B}
  \DeclareMathSymbol{\beta}{\mathalpha}{greek}{"0C}
  \DeclareMathSymbol{\gamma}{\mathalpha}{greek}{"0D}
  \DeclareMathSymbol{\delta}{\mathalpha}{greek}{"0E}
  \DeclareMathSymbol{\epsilon}{\mathalpha}{greek}{"0F}
  \DeclareMathSymbol{\zeta}{\mathalpha}{greek}{"10}
  \DeclareMathSymbol{\eta}{\mathalpha}{greek}{"11}
  \DeclareMathSymbol{\theta}{\mathalpha}{greek}{"12}
  \DeclareMathSymbol{\iota}{\mathalpha}{greek}{"13}
  \DeclareMathSymbol{\kappa}{\mathalpha}{greek}{"14}
  \DeclareMathSymbol{\lambda}{\mathalpha}{greek}{"15}
  \DeclareMathSymbol{\mu}{\mathalpha}{greek}{"16}
  \DeclareMathSymbol{\nu}{\mathalpha}{greek}{"17}
  \DeclareMathSymbol{\xi}{\mathalpha}{greek}{"18}
  \DeclareMathSymbol{\pi}{\mathalpha}{greek}{"19}
  \DeclareMathSymbol{\rho}{\mathalpha}{greek}{"1A}
  \DeclareMathSymbol{\sigma}{\mathalpha}{greek}{"1B}
  \DeclareMathSymbol{\tau}{\mathalpha}{greek}{"1C}
  \DeclareMathSymbol{\upsilon}{\mathalpha}{greek}{"1D}
  \DeclareMathSymbol{\phi}{\mathalpha}{greek}{"1E}
  \DeclareMathSymbol{\chi}{\mathalpha}{greek}{"1F}
  \DeclareMathSymbol{\psi}{\mathalpha}{greek}{"20}
  \DeclareMathSymbol{\omega}{\mathalpha}{greek}{"21}
  \DeclareMathSymbol{\varepsilon}{\mathalpha}{greek}{"22}
  \DeclareMathSymbol{\vartheta}{\mathalpha}{greek}{"23}
  \DeclareMathSymbol{\varpi}{\mathalpha}{greek}{"24}
  \DeclareMathSymbol{\varrho}{\mathalpha}{greek}{"25}
  \DeclareMathSymbol{\varsigma}{\mathalpha}{greek}{"26}
  \DeclareMathSymbol{\varphi}{\mathalpha}{greek}{"27}
  \DeclareSymbolFont{otone}{OT1}{cmr}{m}{n}
  \DeclareMathSymbol{\Gamma}{\mathalpha}{otone}{0}
  \DeclareMathSymbol{\Delta}{\mathalpha}{otone}{1}
  \DeclareMathSymbol{\Theta}{\mathalpha}{otone}{2}
  \DeclareMathSymbol{\Lambda}{\mathalpha}{otone}{3}
  \DeclareMathSymbol{\Xi}{\mathalpha}{otone}{4}
  \DeclareMathSymbol{\Pi}{\mathalpha}{otone}{5}
  \DeclareMathSymbol{\Sigma}{\mathalpha}{otone}{6}
  \DeclareMathSymbol{\Upsilon}{\mathalpha}{otone}{7}
  \DeclareMathSymbol{\Phi}{\mathalpha}{otone}{8}
  \DeclareMathSymbol{\Psi}{\mathalpha}{otone}{9}
  \DeclareMathSymbol{\Omega}{\mathalpha}{otone}{10}
  \DeclareSymbolFont{syms}{OML}{cmm}{m}{it}
  \DeclareMathSymbol{\partial}{\mathord}{syms}{"40}
  \DeclareMathAlphabet{\mathbold}{OML}{cmm}{b}{it}
  \DeclareSymbolFont{largesymbols}{OMX}{cmex}{m}{n}

  \DeclareMathAlphabet{\mathcal}{OMS}{cmsy}{m}{n}

\usepackage{CJKutf8} %
\newcommand{\zh}[1]{\begin{CJK}{UTF8}{gbsn}#1\end{CJK}}
\usepackage{mdframed} %
\usepackage{mfirstuc}
\usepackage{inconsolata}
\usepackage{natbib}
\usepackage{tikz}
\usepackage{subcaption}
\usepackage[many]{tcolorbox}
\usepackage{listings}
\definecolor{codegreen}{rgb}{0,0.6,0}
\definecolor{codegray}{rgb}{0.5,0.5,0.5}
\definecolor{codepurple}{rgb}{0.58,0,0.82}
\definecolor{backcolour}{rgb}{0.95,0.95,0.92}
\lstdefinestyle{mystyle}{
    backgroundcolor=\color{backcolour},
    commentstyle=\color{codegreen},
    keywordstyle=\color{magenta},
    numberstyle=\tiny\color{codegray},
    stringstyle=\color{codepurple},
    basicstyle=\ttfamily\footnotesize,
    breakatwhitespace=false,
    breaklines=true,
    captionpos=b,
    keepspaces=true,
    numbers=left,
    numbersep=5pt,
    showspaces=false,
    showstringspaces=false,
    showtabs=false,
    tabsize=2
}
\newcommand{\aka}{also known as}

\usepackage{textgreek}
\usepackage{upgreek}
\usepackage{algpseudocode}

\newcommand{\colorFiveOneFiveNine}{\textcolor{red}}
\newcommand{\colorFiveOneEight}{\textcolor{blue}}
\newcommand{\colorTwoSixThreeSix}{\textcolor{darkgray}}
\newcommand{\colorTwoNineNineSixTwo}{\textcolor{orange}}
\newcommand{\colorFiveFiveOneNine}{\textcolor{purple}}
\newcommand{\colorFiveTwoSixTwo}{\textcolor{magenta}}
\newcommand{\colorTwoNineNineSixOne}{\textcolor{cyan}}
\newcommand{\colorEightNineNineNine}{\textcolor{brown}}

\usepackage{lipsum} %
\usepackage{ACL2023}

\usepackage{times}
\usepackage{latexsym}

\usepackage{microtype}

\title{Byte BPE Tokenization as an Inverse string Homomorphism}

\author{Saibo Geng, Sankalp Gambhir, Chris Wendler, Robert West \\
EPFL, Switzerland \\
\texttt{saibo.geng@epfl.ch}}

\begin{document}
\maketitle

\begin{abstract}
Tokenization is an important preprocessing step in the training and inference of large language models (LLMs).
While there has been extensive research on the expressive power of the neural achitectures used in LLMs, the impact of tokenization has not been well understood.
In this work, we demonstrate that tokenization, irrespective of the algorithm used, acts as an inverse homomorphism between strings and tokens. 
This suggests that the character space of the source language and the token space of the tokenized language are homomorphic, preserving the structural properties of the source language.
Additionally, we explore the concept of proper tokenization, which refers to an unambiguous tokenization returned from the tokenizer. 
Our analysis reveals that the expressiveness of neural architectures in recognizing context-free languages is not affected by tokenization.

\end{abstract}

\section{Introduction}\label{sec:introduction}

\begin{figure}[t!]
\begin{mdframed} %
\textbf{1. Tokenization is inverse-homomorphic from token IDs to ASCII\footnote{A leading space is omitted before the token \textit{Hello}}}
\begin{align*}
    f_{\text{detok}}(\textcolor{red}{15496}) &= \textup{``\textcolor{red}{Hello}''} \\
    f_{\text{detok}}(\textcolor{blue}{2159}) &= \textup{``\textvisiblespace\textcolor{blue}{World}''}.\\
f_{\text{detok}}([\textcolor{red}{15496}, \textcolor{blue}{2159}]) &= \textup{``Hello\textvisiblespace World ''} 
\end{align*}
\textbf{2. Tokenization is inverse-homomorphic from token IDs to Unicode bytes Tokenization}
\begin{align*}
    f_{\text{detok}}(\textcolor{magenta}{19526}) &= \text{``\zh{\textcolor{magenta}{E4 BD}}''}\\
     f_{\text{detok}}(\textcolor{orange}{254}) &= \text{``\zh{\textcolor{orange}{A0}}''}.\\
    f_{\text{detok}}([\textcolor{magenta}{19526}, \textcolor{orange}{254}]) &=  \text{``E4 BD A0 (\zh{你})''} \\
\end{align*}
\begin{itemize}
    \denselist
    \item $\text{Detokenization} = f_{\text{detok}}:\mathbb{N^*} \to \Sigma^*$
    \item $\text{Tokenization} = f_{\text{detok}}^{-1}:\Sigma^* \to \mathbb{N^*}$
    \item \texttt{E4 BD A0} is the utf-8 encoding of the Chinese character \zh{你}(U+4F60), means \textit{you}.
\end{itemize}
\end{mdframed} %
\captionsetup{type=figure}
\caption{Tokenization and Detokenization example illustrating the homomorphism property in with OpenAI GPT-2's Tokenizer.}
\label{fig:figure1}

\end{figure}

As shown in \Figref{fig:figure_2_tokenizer_output_for_nested_brackets}, tokenizing a few strings with a very simple structure, such as balanced parentheses, results in a non-trivial sequence of token IDs that do not directly correspond to the original characters.
The misalignment between the token IDs and the original characters is further amplified when the strings contain Unicode characters, which are represented by multiple bytes (or multiple token IDs ) as shown in \Figref{fig:figure1}.
In this work, we aim to answer the following question:

Q1: Given a context-free or regular language $L$ over the character alphabet, what is the structure of the token language $L^\prime$ after tokenization?

This question natrually arises when we want to study the recognition power of language models (LLMs) for a certain category, such as context-free languages, by writing a context-free grammar in the character space and feeding it directly to the LLM without worrying about the structure being lost after tokenization.
To answer this question, we first formalize the tokenization process as a mapping from the character alphabet to the token ID alphabet.
Counterintuitively, we first show that the tokenization is not an homomorphic mapping, \ie, it does not preserve the structure of the input string language.
However, we show that the inverse of the tokenization process, which we refer to as detokenization, is an homomorphic mapping.
This homomorphic property of detokenization allows us to establish a connection between the token language $L^\prime$ and the original string language $L$.
We thus obtain a preliminary answer to Q1: the token language $L^\prime$ retains the structure of the original string language $L$. 
If the original language $L$ is a context-free (regular) language, then the token language $L^\prime$ is also a context-free (regular) language.
We then extend our discussion to the following question:

Q2: Is the answer to Q1 affected by the presence of Unicode characters in the source language?

We show that the above homomorphic property of detokenization breaks down when the source language contains Unicode characters.
This is because a single Unicode character can be represented by multiple token IDs. 
However, a closer look reveals that the tokenization process is actually operating on byte-level tokenizations of the Unicode characters.
We show that the support for Unicode characters can be naturally integrated into the homomorphic framework by transforming the grammar with Unicode characters to a new grammar with byte-level alphabet.
We thus obtain the answer to Q2: the presence of Unicode characters does not affect the homomorphic property of tokenization.
Finally, we extend our discussion to the following question:

Q3: The tokenization mentioned above is actually a one-to-many mapping. But in practice, the tokenizer selects one of the possible tokenizations based on some rules. How does this affect the structure of the token language?

We refer to the special tokenization selected by the tokenizer as the \textit{proper tokenization}.
The actual \textit{proper tokenization language} is a special subset of the \textit{tokenization language}.
We provide some insights into the structure of the \textit{proper tokenization language} while not providing a complete answer to Q3.

\begin{figure*}[ht]
\centering
\small
\begin{tabular}{llll}
\hline
\textbf{Depth} & \textbf{String} & \textbf{Tokenization}  & \textbf{Tokens} \\
\hline
0 & \textbf{""}              & [ 1  ] &  BOS = 1 \\
1 & \textbf{"\colorFiveOneFiveNine{[]}"}            & [ 1, \colorFiveOneFiveNine{5159}  ] &  \textbf{\colorFiveOneEight{\textvisiblespace [}} = \colorFiveOneEight{ 518} \\
2 & \textbf{"\colorFiveOneEight{[}\colorTwoSixThreeSix{[]}\colorTwoNineNineSixTwo{]}"}          & [ 1, \colorFiveOneEight{518}, \colorTwoSixThreeSix{2636}, \colorTwoNineNineSixTwo{29962} ] &  \textbf{\colorTwoSixThreeSix{[]}} = \colorTwoSixThreeSix{2636} \\
3 & \textbf{"\colorFiveFiveOneNine{[[}\colorTwoSixThreeSix{[]}\colorFiveTwoSixTwo{]]}"}        & [ 1, \colorFiveFiveOneNine{5519}, \colorTwoSixThreeSix{2636}, \colorFiveTwoSixTwo{5262} ] &  \textbf{\colorFiveFiveOneNine{\textvisiblespace[[}} = \colorFiveFiveOneNine{  5519} \\
4 & \textbf{"\colorFiveFiveOneNine{[[}\colorTwoNineNineSixOne{[[}\colorTwoSixThreeSix{[]}\colorFiveTwoSixTwo{]]}\colorTwoNineNineSixTwo{]}"}      & [ 1, \colorFiveFiveOneNine{5519}, \colorTwoNineNineSixOne{29961}, \colorTwoSixThreeSix{2636}, \colorFiveTwoSixTwo{5262}, \colorTwoNineNineSixTwo{29962} ] &  \textbf{\colorTwoNineNineSixOne{[[}} = \colorTwoNineNineSixOne{29961} \\
5 & \textbf{"\colorFiveFiveOneNine{[[}\colorEightNineNineNine{[[[}\colorTwoSixThreeSix{[]}\colorFiveTwoSixTwo{]]}\colorFiveTwoSixTwo{]]}"}    & [ 1, \colorFiveFiveOneNine{5519}, \colorEightNineNineNine{8999}, \colorTwoSixThreeSix{2636}, \colorFiveTwoSixTwo{5262}, \colorFiveTwoSixTwo{5262} ] &  \textbf{\colorEightNineNineNine{[[[}} = \colorEightNineNineNine{8999} \\
6 & \textbf{"\colorFiveFiveOneNine{[[}\colorEightNineNineNine{[[[}\colorTwoNineNineSixOne{[[}\colorTwoSixThreeSix{[]}\colorFiveTwoSixTwo{]]}\colorFiveTwoSixTwo{]]}\colorTwoNineNineSixTwo{]}"}  & [ 1, \colorFiveFiveOneNine{5519}, \colorEightNineNineNine{8999}, \colorTwoNineNineSixOne{29961}, \colorTwoSixThreeSix{2636}, \colorFiveTwoSixTwo{5262}, \colorFiveTwoSixTwo{5262}, \colorTwoNineNineSixTwo{29962} ] &  \textbf{\colorTwoNineNineSixTwo{]}} = \colorTwoNineNineSixTwo{29962} \\
7 & \textbf{"\colorFiveFiveOneNine{[[}\colorEightNineNineNine{[[[}\colorEightNineNineNine{[[[}\colorTwoSixThreeSix{[]}\colorFiveTwoSixTwo{]]}\colorFiveTwoSixTwo{]]}\colorFiveTwoSixTwo{]]}"}& [ 1, \colorFiveFiveOneNine{5519}, \colorEightNineNineNine{8999}, \colorEightNineNineNine{8999}, \colorTwoSixThreeSix{2636}, \colorFiveTwoSixTwo{5262}, \colorFiveTwoSixTwo{5262}, \colorFiveTwoSixTwo{5262} ] &  \textbf{\colorFiveTwoSixTwo{]]}} = \colorFiveTwoSixTwo{5262} \\
8 & \textbf{"\colorFiveFiveOneNine{[[}\colorEightNineNineNine{[[[}\colorEightNineNineNine{[[[}\colorTwoNineNineSixOne{[[}\colorTwoSixThreeSix{[]}\colorFiveTwoSixTwo{]]}\colorFiveTwoSixTwo{]]}\colorFiveTwoSixTwo{]]}\colorTwoNineNineSixTwo{]}"}& [ 1, \colorFiveFiveOneNine{5519}, \colorEightNineNineNine{8999}, \colorEightNineNineNine{8999}, \colorTwoNineNineSixOne{29961}, \colorTwoSixThreeSix{2636}, \colorFiveTwoSixTwo{5262}, \colorFiveTwoSixTwo{5262}, \colorFiveTwoSixTwo{5262}, \colorTwoNineNineSixTwo{29962} ] &  \textbf{\colorFiveOneFiveNine{  \textvisiblespace []}} = \colorFiveOneFiveNine{ 5159} \\
\hline
\end{tabular}
\caption{
Tokenization Output for Nested Brackets Using LLaMA-2 Tokenizer
}
\label{fig:figure_2_tokenizer_output_for_nested_brackets}
\end{figure*}

\subsection*{Contributions}\label{subsec:contributions}

\begin{itemize}
    \item \textbf{Tokenization Formalization:} We formalize tokenization as a mapping between character and token ID alphabets. While tokenization itself is not homomorphic, we show that its inverse, detokenization, preserves the structure of the original string language.
    \item \textbf{Language Preservation:} We demonstrate that token languages retain the structure of the original context-free or regular languages, answering our first research question (Q1).
    \item \textbf{Unicode Integration:} We extend the homomorphic framework to handle Unicode characters by treating tokenization at the byte level, ensuring that Unicode support does not affect the structure of the token language (Q2).
    \item \textbf{Proper Tokenization:} We introduce \textit{proper tokenization} and analyze some of its structural properties, providing insights into the structure of the proper tokenization language (Q3).
\end{itemize}

These contributions offer a framework to understand tokenization’s role in preserving language structure in large models.

\section{Preliminaries}\label{sec:background}

\subsection{Context-free grammar and language}\label{subsec:context-free-language}

\begin{definition}[Context-free Grammar]\label{def:cfg}
    A \emph{context-free grammar (CFG)} is a 4-tuple $G = (V, \Sigma, P, S)$, where
    \begin{itemize}
        \item $V$ is a finite set of non-terminal symbols (variables),
        \item $\Sigma$ is a finite set of terminal symbols,
        \item $P$ is a finite set of production rules, each of the form $A \rightarrow \alpha$, where $A \in N$ and $\alpha \in (N \cup \Sigma)^*$,
        \item $S \in N$ is the start symbol.
    \end{itemize}
\end{definition}

\begin{definition}[Formal Language]\label{def:formal-language}
    A \emph{formal language} $L$ is a set of strings over an alphabet $\Sigma$, where a string is a finite sequence of symbols from $\Sigma$.
\end{definition}

If $G(V, \Sigma, P, S)$ is a CFG, the language of $G$, denoted $L(G)$, is the set of all strings of terminal symbols that can be derived from the start symbol $S$.
If a language $L$ is the language of some CFG, then $L$ is called a \emph{context-free language(CFL)}.

\begin{definition}[Pushdown Automaton]\label{def:pda}
    A \emph{pushdown automaton (PDA)} is a 7-tuple $M = (Q, \Sigma, \Gamma, \delta, q_0, Z_0, F)$, where
    \begin{itemize}
        \item $Q$ is a finite set of states,
        \item $\Sigma$ is a finite set of input symbols,
        \item $\Gamma$ is a finite set of stack symbols,
        \item $\delta: Q \times (\Sigma \cup \{\epsilon\}) \times \Gamma \rightarrow 2^{Q \times \Gamma^*}$ is the transition function,
        \item $q_0 \in Q$ is the start state,
        \item $Z_0 \in \Gamma$ is the initial stack symbol,
        \item $F \subseteq Q$ is the set of accepting states.
    \end{itemize}
\end{definition}

\begin{theorem}[Pushdown Automaton and Context-free Grammar]\label{thm:pda-cfg}
    For every context-free grammar $G$, there exists a pushdown automaton $M$ that accepts the language $L(G)$.
\end{theorem}

\Thmref{thm:pda-cfg} implies that one can always construct a PDA to decide whether a given string belongs to a context-free language.

\begin{definition}[String Homomorphism]\label{def:homomorphism}
    Given two operations $\oplus$ and $\odot$ on two alphabets $\Sigma^*$ and $\mathbb{N}^*$ respectively, a function $h: \Sigma^* \rightarrow T^*$ is a string homomorphism if $\forall u, v \in \Sigma^*, h(u \oplus v) = h(u) \odot h(v)$.
\end{definition}

In the following, we assume that $\oplus$ and $\odot$ are both string concatenation operations and use $xy$ to denote the concatenation of two elements $x$ and $y$.
Thus, a mapping $h$ is a string homomorphism if it preserves the concatenation of strings.
One can apply a homomorphism to a language $L$ by applying it to each string in the language, which results in a new language $h(L)$.
That is,  $h(L) = \{h(w) \mid w \in L\}$ is the image of $L$ under $h$.

\begin{definition}[Inverse Homomorphism]
Given a string homomorphism $h: \Sigma^* \rightarrow \mathbb{N}^*$, the inverse function $h^{-1}: \mathbb{N}^* \rightarrow \Sigma^*$ is called an inverse homomorphism.
\end{definition}

The inverse homomorphism $h^{-1}(L)$ includes all strings in $\Sigma^*$ that map to strings in $L$ under $h$.

\begin{theorem}[Closure under Inverse Homomorphism]
\label{theorem:inverseClosure}
    If $L$ is a context-free(regular) language and $h: \Sigma^* \rightarrow \mathbb{N}^*$ is a homomorphism, then the inverse homomorphic image $h^{-1}(L)$ is also a context-free(regular) language~\citep[Theorem 7.30]{book_intro_to_automata_languages_computation}.
\end{theorem}

\begin{theorem}[Closure under intersection]
\label{theorem:intersection_of_regular_and_context_free}
    If $L_1$ is a context-free(regular) language and $L_2$ is a regular language, then the intersection $L_1 \cap L_2$ is also a context-free(regular) language ~\citep[Theorem 7.27]{book_intro_to_automata_languages_computation}.
\end{theorem}

\subsection{Tokenization}\label{subsec:tokenization_encoding}

In the context of LLM, we have two alphabets:
\begin{enumerate}
    \item the character alphabet $\Sigma$ which is typically a charset, \ie Unicode characters or ASCII.
    \item the token ID alphabet $\mathbb{N}$ which is the set of all possible token IDs in a language model's vocabulary, \ie $\mathbb{N} = \{0, 1, \ldots, |V| - 1\}$ where $V$ is the vocabulary of the language model's tokenizer.
\end{enumerate}

\begin{definition}[Tokenization]
The tokenization function \footnote{In some literature, \textit{tokenization} refers specifically to the process of splitting text, while the term \textit{encoding} is used for the mapping from tokens to IDs. In this work, we use tokenization to cover both processes.} $f_{\text{tok}}: \Sigma^* \rightarrow \mathbb{N}^*$ maps a string to a sequence of sub-word units, or tokens, which are indexed by token IDs and can be fed into a model. 
\end{definition}
The function $f_{\text{tok}}$ is injective but not \textbf{surjective}, since not every sequence in $\mathbb{N}^*$ corresponds to a string in $\Sigma^*$.
\begin{definition}[Detokenization]
The detokenization function $f_{\text{detok}}: \mathbb{N}^* \rightarrow \Sigma^*$ does the opposite of tokenization. It reconstructs the original string by converting the token IDs back into their respective sub-word tokens and concatenating them. 
\end{definition}

By construct, $f_{\text{detok}}(f_{\text{tok}}(x)) = x$ for all $x \in \Sigma^*$.

\begin{definition}[Extended Tokenization]
The extended tokenization function $F_{\text{tok}}: \Sigma^* \rightarrow \mathbb{N}^*$ is a \textbf{surjective} extension of $f_{\text{tok}}$, mapping a string to all possible valid tokenizations. 
\end{definition}
An illustration of the difference between a proper tokenization and an extended tokenization is shown in Listing~\ref{lst:decode_non_injective}.

\begin{proposition}[]
The detokenization function $f_{\text{detok}}$ is the inverse of the tokenization function $F_{\text{tok}}$. \label{proposition:detokenization_is_inverse_of_tokenization}
\end{proposition}

\noindent Given a string $s \in \Sigma^*$, we can distinguish between three types of tokenizations:
\begin{enumerate}
\denselist
    \item $f_{\text{tok}}(s)$ as the \textbf{proper tokenization} of $s$
    \item $F_{\text{tok}}(s)$ as the \textbf{extended tokenization} of $s$
    \item $F_{\text{tok}}(s) \setminus f_{\text{tok}}(s)$ as the \textbf{improper tokenization} of $s$
\end{enumerate}
In practice, \textbf{Proper tokenization} is the unique tokenization of a string that is directly returned by the tokenizer.
As $f_{\text{tok}}$ is not surjective, we define the image of $f_{\text{tok}}$, which is a strict subset of $\mathbb{N}^*$, as the \textbf{proper tokenization space} $\mathbb{N}_{\text{proper}}^*$.

\begin{lstlisting}[language=Python, numbers=none, caption=The text "aaaa" has more than one tokenizations while only the first one is the proper tokenization.,label={lst:decode_non_injective}]
from transformers import GPT2Tokenizer 
tokenizer = GPT2Tokenizer.from_pretrained("gpt2")                       

tokenizer.encode("a") #[64]                                                   

tokenizer.encode("aa") #[7252]                                                 

tokenizer.encode("aaa") #[46071]                                                

tokenizer.encode("aaaa")  #[24794]                                              

tokenizer.decode([24794]) #'aaaa'                                               

tokenizer.decode([46071, 64]) #'aaaa'                                           

tokenizer.decode([7252,7252]) #'aaaa'                                          

tokenizer.decode([7252,64,64]) #'aaaa'                                        

tokenizer.decode([64,7252,64])  #'aaaa'                                       

tokenizer.decode([64,64,64,64]) #'aaaa'                                       

\end{lstlisting}

\xhdr{Unicode Support}
Byte-level tokenization\footnote{\aka{} byte-level encoding}~\citep{wang2019neural_bbpe, Radford2019LanguageMA_gpt2} is the standard way to provide support for Unicode characters.
Unlike traditional character-level tokenization, byte-level tokenization first converts the text into a byte sequence according to a format such as UTF-8, and then applies tokenization to the byte sequence.
The resulting token IDs are chunks of bytes instead of characters, which allows the model to support effectively any Unicode character.

\subsection{Tokenization languages}\label{sec:tokenization_languages}

Given a formal language $L$ defined on an alphabet $\Sigma$ (either ASCII or Unicode), with a context-free grammar $G$ generating $L$, we investigate the structure of the token language $L^\prime$ after tokenization.

Aligned with the terms used in the previous sections, we define the following terms:

\begin{itemize}
\denselist
    \item \textit{Source language} $L$ is a context-free or regular language over the character alphabet $\Sigma$.
    \item \textit{Extended tokenization language} $L^\prime_E$ is the set of all possible tokenizations of strings in $L$, i.e. the image of $L$ under the extended tokenization $F_{\text{tok}}$.
    \item \textit{Proper tokenization language} $L^{\prime}$ is the set of tokenizations returned by the tokenizer, i.e. the image of $L$ under the proper tokenization $f_{\text{tok}}$.
    \item \textit{Improper tokenization language} $L^{\prime}_I$ is the set of tokenizations that are not returned by the tokenizer, i.e. $L^{\prime}_I = L^\prime_E \setminus L^{\prime}$.
\end{itemize}

\section{Extended Tokenization preserving Language Structure}\label{sec:structure_and_decidability}

In this section, we study Q1: Given a context-free or regular language $L$ over the character alphabet, what is the structure of the token language $L^\prime$ after tokenization?
We start with showing that \textbf{extended tokenization} can be seen as an inverse homomorphism from $\Sigma^*$ to $\mathbb{N}^*$, which implies that the structure of the source language is preserved.
At the end, we show a construction of a pushdown automaton (PDA) that recognizes the token language.

\subsection{Extended Tokenization is inverse homomorphism}

\begin{proposition}\label{proposition:tokenization_is_not_homomorphism}
    Tokenization function $f_{\text{tok}}$ is \textbf{not} homomorphic from $\Sigma^*$ to $\mathbb{N}^*$ under the concatenation operation.
\end{proposition}
\vspace{1em} %
One can easily verify this by considering the following counter-example.
\begin{example}
    With GPT-3 tokenizer, the brackets  \texttt{[} and  \texttt{]} are individually tokenized as $58$ and $60$, respectively, but the combined  \texttt{[]} is tokenized as $21737$.
\end{example}
\vspace{1em} %
In contrast, the detokenization function is homomorphic, \ie $F_{\text{detok}}([n_1, n_2]) = [F_{\text{detok}}(n_1), F_{\text{detok}}(n_2)], \forall n_1, n_2 \in \mathbb{N}^*$ as shown in \Figref{fig:figure1}
This is not surprising, as the detokenization function, as the name suggests, performs the following steps:

\begin{enumerate}
    \denselist
    \item Mapping token IDs back to their corresponding tokens.
    \item Concatenating these tokens to reconstruct the string.
    \item Performing necessary post-processing to restore the original string format.
\end{enumerate}

We can now state the following proposition:
\begin{proposition}\label{proposition:detokenization_is_inverse_of_tokenization}
    The detokenization function $F_{\text{detok}}: \mathbb{N}^* \rightarrow \Sigma^*$ is homomorphic under the concatenation operation.
\end{proposition}
\vspace{1em} %
As a direct consequence of the Proposition~\ref{proposition:detokenization_is_inverse_of_tokenization}, we have the following corollary:
\begin{corollary}
    The extended tokenization function $F_{\text{tok}}$ is an \textbf{inverse homomorphism} from $\Sigma^*$ to $\mathbb{N}^*$ under the concatenation operation.
\end{corollary}
\vspace{1em} %
All major tokenization schemes, including Byte Pair Encoding (BPE)~\citep{bpe_sennrich-etal-2016-neural}, WordPiece, and SentencePiece~\citep{kudo2018sentencepiece}, exhibit this homomorphic property in their detokenization functions.

Using the closure properties of context-free languages under inverse homomorphism (Theorem~\ref{theorem:inverseClosure}), we can state the following proposition:

\begin{proposition}\label{proposition:token_language_is_context_free}
 The extended token language $L^\prime_{E}\subseteq \mathbb{N}^*$ is a context-free language if the original language $L\subseteq \Sigma^*$ is a context-free language.
\end{proposition}
\vspace{1em} %

\subsection{Token-space automata construction}\label{subsec:construction}

\begin{figure}[ht]
    \centering
    \includegraphics[width=1.0\linewidth]{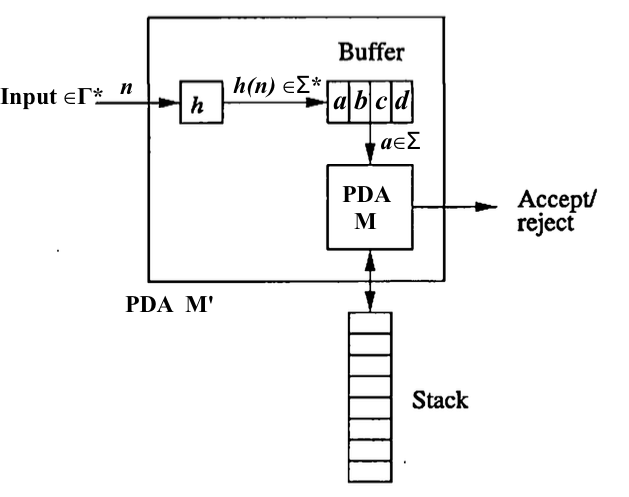}
    \caption{\textbf{Construction of a PDA $M^\prime$ to accept language $h^{-1}(L)$}. In the context of LLM, the input $a$ is a token ID, the homomorphism $h$ is \textit{detokenization}, the \textit{buffer} is used to store the token $h(a)$, the \textit{PDA state} is the current state of the PDA in the character space, and the \textit{PDA stack} is the stack of the PDA in the character space.
    }
    \label{fig:PDA_simulation}
\end{figure}

In this section, we explain how to construct a recognizer for the extended token language $L^{\prime}_E$ based on the recognizer for the string language $L$.
The main idea is based on the construction of a pushdown automaton (PDA) for the inverse homomorphism of a context-free language sketched in~\citet[Theorem 7.30]{book_intro_to_automata_languages_computation}.
Given a homomorphism $h$ from alphabet $\mathbb{N}$ to alphabet $\Sigma$, and $L$ being a context-free language over $\Sigma$, the construction of a PDA to accept language $L^\prime = h^{-1}(L)$ is shown in~\Figref{fig:PDA_simulation}.
As stated in~\Thmref{thm:pda-cfg}, we can always construct a PDA $M$ which reads the input string in the alphabet $\Sigma$ and accepts the language $L$.
The construction of such a PDA is standard and well-known in the literature~\citep[Chap 6.3.1]{book_intro_to_automata_languages_computation}.
We then construct a PDA $M^\prime$ which reads the input string in the alphabet $\mathbb{N}$ (token IDs in our case) and accepts the language $L^\prime_E=F_{\text{tok}}(L)$.
The working of the PDA $M^\prime$ is as follows:
\begin{enumerate}
    \item It applies the homomorphism $h$(detokenization in our case) to the input token ID $a$ and puts the result $h(a)$ into the buffer, \ie mapping the token ID to the corresponding string in the character space.
    \item The underlying PDA $M$ in the character space reads the input characters $h(a)$ and updates its state and stack accordingly.
\end{enumerate}
The resulting PDA $M^\prime$ reads the token IDs as input and decides whether the token IDs form a valid string in the token language $tok(L)$.

\section{Tokenization with Unicode characters}\label{sec:construct-gcd}

In this section, we answer the Q2: Is the presence of Unicode characters in the input language a challenge for the structure-preserving property of tokenization?

When the source langauge contains Unicode characters, the tokenization process becomes more complex because a single character can be represented by multiple tokens which are not detokenizable independently.
For example, the Chinese character \zh{你}(U+4F60, \textit{you}) is tokenized as $[19526, 254]$ in the GPT-2 tokenizer but the token $19526$ or $254$ alone does not correspond to any character.
Knowing only the token $19526$ is insufficient to determine the character \zh{你}, as the context provided by the token $254$ is also necessary, as illustrated in \Figref{fig:figure1}.
This dependency of the next token seems breaking the homomorphic property of the detokenization function as shown in~\Figref{fig:figure1}.
However, considering that the tokenization function is actually operating on byte-level tokenizations of the Unicode characters, we can prove that the token language still retains the structure of the original language.

\begin{proposition}\label{prop:byte_level_inverse_homomorphism}
    Byte-level tokenization is inverse-homomorphic from token IDs to byte sequences.
\end{proposition}

With the exactly same argument as in Proposition~\ref{proposition:detokenization_is_inverse_of_tokenization}, we can show that the byte-level tokenization is inverse-homomorphic from token IDs to byte sequences.
The character-level tokenization on ASCII characters is a special case of byte-level tokenization where each character is represented by a single byte.

\begin{lemma}[Character Encoding Scheme Closure]\label{lemma:context-free-closure-under-utf-8-encoding}
    Context-free languages are closed under any finite-length character encoding scheme such as UTF-8 or UTF-16, which maps characters to byte sequences.
\end{lemma}

Lemma~\ref{lemma:context-free-closure-under-utf-8-encoding} is trivially true because a finite-length character encoding scheme is a string replacement operation.

Combining Proposition~\ref{prop:byte_level_inverse_homomorphism} and Lemma~\ref{lemma:context-free-closure-under-utf-8-encoding}, we can show that the following three langauges all have the same structure:
\begin{itemize}
    \item The source language $L$ over the character alphabet.
    \item The byte-level language $L_b$, which is the image of $L$ under the character encoding scheme such as UTF-8.
    \item The token language $L_t$, which is the image of $L_b$ under the byte-level tokenization function.
\end{itemize}

\section{Proper tokenization language}\label{sec:closure_under_proper_tokenization}
In the previous section, we have shown that the image of a context-free language under extended tokenization, either byte-level or character-level, is still a context-free language.
In this section, we investigate Q3: Given a context-free or regular source language $L$, what is the structure of the proper tokenization language $L^\prime$ after tokenization?

\begin{table}[h]
    \centering
    \small
    \begin{tabular}{p{0.18\linewidth}p{0.72\linewidth}}  %
    \toprule
    \textbf{Model} & \textbf{BPE} \\
    \midrule
    \textbf{Training} & Starts from a basic token vocabulary, i.e. 256 byte values as the initial tokens, and learns rules to merge tokens \\
    \midrule
    \textbf{Learns} & An ordered list of binary merges and a vocabulary of final tokens \\
    \midrule
    \textbf{Tokenizing} & Given a text, maps it to a sequence of bytes and iteratively applies the merge rules (in order) to the byte sequence until no more merges can be performed \\
    \bottomrule
    \end{tabular}
    \caption{Byte Pair Encoding (BPE) Algorithm}
    \label{tab:bpe_algorithm}
\end{table}

\subsection{Byte Pair Encoding (BPE)}\label{sec:bpe_algorithm}

As described in \Tabref{tab:bpe_algorithm}, the BPE algorithm iteratively applies the merge operation (in order) to the token sequence until no more merge operations can be performed.
The token sequence after the last merge operation is the proper tokenization of the input string.

\xhdrNoPeriod{Two types of improper tokenization:}
We define two types of improper tokenization for BPE:
\begin{itemize}
    \denselist
    \item \textbf{Mergeable tokenization}: A tokenization \( t_1, t_2, \cdots, t_n \) is called mergeable if there exists a token \( t_{i+1} \) that can be merged with \( t_i \).
    \item \textbf{Tokenization with wrong merge order}: A tokenization \( t_1, t_2, \cdots, t_n \) is called a tokenization with the wrong merge order if it is not the proper tokenization and can not be merged further.
\end{itemize}

\begin{example}
    Consider a learnt BPE tokenizer with vocabulary $\{a, b, aa, aaa, bb\}$ and the following merge operations in order:
    \begin{enumerate}
        \item $a, a \rightarrow aa$
        \item $aa, a \rightarrow aaa$
        \item $a, b \rightarrow ab$
        \item $b, b \rightarrow bb$
    \end{enumerate}
    Given the input string $aaabb$, the proper tokenization is $aaa, bb$.
    However, if we exchange the order of merge rule 2 and 3, we get the tokenization $aa, ab, b$, which is an \textbf{improper tokenization with wrong merge order}.
\end{example}

It's straightforward to detect mergeable tokens in a token sequence by looping through the pairs of tokens and checking if they can be merged.
To detect a tokenization with the wrong merge order, one can use the \textbf{unmerge-remerge} method:
\begin{enumerate}
    \denselist
    \item \textbf{Unmerge}: Given a token sequence $t_1, t_2, \cdots, t_n$, unmerge the tokens in the reverse order of the merge operation to get the sequence of bytes $b_1, b_2, \cdots, b_m$.
    \item \textbf{Remerge}: Apply the BPE algorithm to $b_1, b_2, \cdots, b_m$ to get the proper tokenization $s_1, s_2, \cdots, s_k$.
    \item \textbf{Check}: If $t_1, t_2, \cdots, t_n \neq s_1, s_2, \cdots, s_k$, then $t_1, t_2, \cdots, t_n$ is an improper tokenization with the wrong merge order.
\end{enumerate}

\subsection{Is Proper Tokenization Language Context-Free?}

\begin{proposition}
    Proper tokenization language $L^{\prime}$ is a subset of the Extended tokenization language $L^{\prime}_E$.
    More specifically, it is the \textbf{intersection} of $L^{\prime}_E$ and the proper tokenization space $\mathbb{N}_{\text{proper}}^*$
\end{proposition}

A subset of a context-free language is not necessarily a context-free language. 
However, if we can show that the proper tokenization space $\mathbb{N}_{\text{proper}}^*$ is a regular language, then we can use Theorem~\ref{theorem:intersection_of_regular_and_context_free} to conclude that the proper tokenization language $L^{\prime}$ is also a context-free language.

Due to the closure properties of regular languages under complementation, it is sufficient to show that the improper tokenization language $L^{\prime}_I$ is a regular language.
Due to the closure properties of regular languages under finite union, it is sufficient to show that both mergeable tokenization and tokenization with the wrong merge order are regular languages.
While the first is straightforward, the second is more challenging.

The \textbf{unmerge-remerge} method is a multi-pass algorithm that is not compatible with the finite-state automaton (FSA) construction.

We leave this as an open problem for future study.

\section{Discussion}\label{sec:discussion}

Suppose that we train a language model $M$ as a \textbf{acceptor} to recognize a context-free language $L$.
There are two tokenization languages:
\begin{itemize}
    \item \textbf{Proper tokenization language} $L^\prime$
    \item \textbf{Extended tokenization language} $L^{\prime}_E$
\end{itemize}

Which one does the model $M$ learn?
It's actually sufficient for the model to learn the \textbf{extended tokenization language} $L^{\prime}_E$.
Given two strings $s_1 \in L$ and $s_2 \notin L$, the model $M$ can differentiate between them by checking if the tokenization is in $L^{\prime}_E$ or not.
We conclude that learning the \textbf{extended tokenization language} is sufficient for the model to recognize the context-free language $L$ correctly.
Therefore, the expressiveness of the neural network model is not limited by the tokenization algorithm used.

\section{Related Work}\label{sec:related_work}

\paragraph{Study of Tokenization}

\citet{kudo_subword_2018_unigram} introduced the concept of \textit{proper tokenization} versus general \textit{tokenization}, referring to the ambiguity present in tokenization algorithms. They suggested that this ambiguity could be leveraged during training to improve language model robustness. \citet{singh_tokenization_2024} investigated how different tokenization schemes affect arithmetic tasks in large language models. In their analysis of subword tokenization methods, \citet{bostrom_byte_2020} found that unigram tokenization aligns better with morphological structures and often surpasses byte-pair encoding (BPE) in downstream tasks. Further, \citet{zouhar-etal-2023-tokenization} examined the link between tokenization and channel efficiency, proposing that an efficient tokenizer maximizes the channel's usage from an information-theoretic perspective.

\paragraph{Enforcing Output Structure in LLMs}

Another area of research focuses on constraining the outputs of large language models (LLMs) to adhere to specific grammatical structures, enhancing performance in tasks like code synthesis and semantic parsing.
\citet{deutsch-etal-2019-general} introduced a method to constrain language model outputs using pushdown automata, suitable for generating context-free languages. \citet{kuchnik2023validating_relm} and \citet{willard2023efficientguidedgenerationlarge} explored techniques for restricting LLM outputs to regular languages, while \citet{shin-etal-2021-constrained}, \citet{geng2024grammarconstrained}, and \citet{NEURIPS2023_cd40d0d6} proposed methods for constraining outputs to context-free grammars. Our work, in contrast, focuses on understanding the properties of tokenization algorithms and the structure of the resulting token language. Additionally, regarding \textit{proper tokenization}, \citet{guidance-ai_guidance} introduced \textit{Token Healing}, a method designed to iteratively retokenize input strings, aiming to recover the correct tokenization.

\section{Conclusion}\label{sec:conclusion}

In conclusion, our work formalizes the tokenization process as an inverse homomorphism for context-free languages, providing a rigorous framework for understanding its structural properties. We demonstrated that the structure of context-free and regular languages is preserved through detokenization, ensuring that the expressiveness of neural architectures is not compromised by tokenization. This property holds even in the presence of Unicode characters, with byte-level tokenization allowing for the preservation of language structure. Moreover, we introduced the concept of proper tokenization and highlighted its implications on the structure of the resulting token languages.

Our findings underscore the importance of tokenization as more than a mere preprocessing step; it is a structural operation that affects the language processing capabilities of large language models. Future work could address the complexity of improper tokenizations, especially in practical implementations, and explore the implications of these findings for optimizing language models further.

\bibliography{custom}
\bibliographystyle{acl_natbib}

\appendix

\section{Example of Homomorphic Tokenization API}\label{sec:huggingface-tokenizer-api}
In this section, we investigate the implementation of tokenization in real-world and show that they still preserve the context-free property of the source language.

Recall that a function $f: \Sigma^* \rightarrow \Gamma^*$ is homomorphic if $f(x\oplus y) = f(x)\oplus f(y)$ for any $x,y\in\Sigma^*$.
In the context of LM, we want to know whether the decoding function \texttt{def tokenizer\_decode(token\_ids: List[int]) -> str:}
is homomorphic.
In the following, we will use the API of the \lstinline[language=Python]{tokenizers} library\footnote{\url{https://github.com/huggingface/tokenizers}} to illustrate the tokenization process.
Generally speaking, the decoding function consists of two steps:
\begin{enumerate}
    \item convert the token ids to tokens. \texttt{tokenizer.convert\_ids\_to\_tokens(token\_ids:List[int])-> List[str]}
    \item join the tokens to form a string and apply some post-processing if needed. \texttt{tokenizer.convert\_tokens\_to\_string(tokens:List[str]-> str)}
\end{enumerate}
We will show that the step (2) can cause the homomorphism to break.

\section{Leading space in tokenization}\label{sec:leading-space-in-tokenization}
Many tokenizers, including LLaMA, T5 employ a longstanding practice of distinguishing between prefix token and non-prefix token by baking the space character into the prefix token.
This heuristic breaks the homomorphism because the leading space in the token will be lost if the token is at the beginning of a string.
An example of Hello World tokenized by T5 is given below:

``Hello World'' is tokenized as \texttt{[22172, 3186]} \texttt{[``\textvisiblespace Hello'', ``\textvisiblespace World'']} by LLAMA.

We define \(h\) as the detokenization function and \(h^{-1}\) as the tokenization function:
Given
\begin{align*}
    h(22172) &= ``\text{\textvisiblespace}Hello'', \\
h(3186) &= ``\text{\textvisiblespace}World''. \\
\end{align*}

We see that the homomorphism is broken:

\begin{align*}
h(22172, 3186) &= ``Hello\text{\textvisiblespace}World'' \\
&\neq \\
h(22172) + h(3186) & =  ``\text{\textvisiblespace}Hello\text{\textvisiblespace}World'' \\
\end{align*}

And if we reverse the order of the tokens, we still get the same problem:
\begin{align*}
h(3186, 22172) &= ``World\text{\textvisiblespace}Hello''\\
&\neq \\
h(3186) + h(22172) &=``\text{\textvisiblespace}World\text{\textvisiblespace}Hello''\\
\end{align*}

The above example shows that the tokenization process is not homomorphic and depends on the \textbf{context} of the token in the string, \ie whether the token is at the beginning of the string or not.

However, this is break is relatively easy to fix by simply considering an intermediate CFL, \ie the language with a leading space.

As the operation of adding a leading space to a string is a regular operation, we still get CFL.

\end{document}